%% file: onlineseg_mm2021.tex
\documentclass[sigconf]{acmart}
\usepackage{url}            
\usepackage{booktabs}       
\usepackage{amsfonts}       
\usepackage{nicefrac}       
\usepackage{microtype}      
\usepackage{multirow}
\usepackage{color}
\usepackage{comment}
\usepackage{graphicx}
\usepackage{subfigure}
\usepackage{lipsum}
\usepackage{pifont}
\usepackage{bm}
\usepackage{enumitem}
\usepackage{algorithm}
\usepackage{algpseudocode}
\usepackage{changepage}
\usepackage{array}
\usepackage{booktabs}
\usepackage{xcolor}
\usepackage{url}
\newcommand{\xmark}{\ding{55}}

\AtBeginDocument{%
  \providecommand\BibTeX{{%
    \normalfont B\kern-0.5em{\scshape i\kern-0.25em b}\kern-0.8em\TeX}}}

\copyrightyear{2021}
\acmYear{2021}
\setcopyright{acmcopyright}\acmConference[MM '21]{Proceedings of the 29th ACM
International Conference on Multimedia}{October 20--24, 2021}{Virtual Event, China}
\acmBooktitle{Proceedings of the 29th ACM International Conference on Multimedia
(MM '21), October 20--24, 2021, Virtual Event, China}
\acmPrice{15.00}
\acmDOI{10.1145/3474085.3475443}
\acmISBN{978-1-4503-8651-7/21/10}

\begin{document}
\fancyhead{}
\title{An EM Framework for Online Incremental Learning of \\Semantic Segmentation}

\author{Shipeng Yan$^{1,2,3*}$, Jiale Zhou$^{1*}$, Jiangwei Xie$^{1}$, Songyang Zhang$^{1}$, Xuming He$^{1,4\dagger}$}
\affiliation{
	\institution{$^{1}$ShanghaiTech University}
	\institution{$^{2}$Shanghai Institute of Microsystem and Information Technology, Chinese Academy of Sciences}
	\institution{$^{3}$University of Chinese Academy of Sciences}
	\institution{$^{4}$Shanghai Engineering Research Center of Intelligent Vision and Imaging}
	\institution{\{yanshp, zhoujl, xiejw, zhangsy1, hexm\}@shanghaitech.edu.cn} \city{} \country{}
}

\thanks{$*$Both authors contributed equally. $\dagger$: Corresponding author. This work was supported by Shanghai Science and Technology Program 21010502700. }



\begin{abstract}
Incremental learning of semantic segmentation has emerged as a promising strategy for visual scene interpretation in the open-world setting. However, it remains challenging to acquire novel classes in an online fashion for the segmentation task, mainly due to its continuously-evolving semantic label space, partial pixelwise ground-truth annotations, and constrained data availability.   
To address this, we propose an incremental learning strategy that can fast adapt deep segmentation models without catastrophic forgetting, using a streaming input data with pixel annotations on the novel classes only. 
To this end, we develop a unified learning strategy based on the Expectation-Maximization (EM) framework, which integrates an iterative relabeling strategy that fills in the missing labels and a rehearsal-based incremental learning step that balances the stability-plasticity of the model. Moreover, our EM algorithm adopts an adaptive sampling method to select informative training data and a class-balancing training strategy in the incremental model updates, both improving the efficacy of model learning. 
We validate our approach on the PASCAL VOC 2012 and ADE20K datasets, and the results demonstrate its superior performance over the existing incremental methods.
\end{abstract}




\maketitle

\input{data/introduction.tex}
\input{data/relatedwork.tex}

\input{data/methods.tex}
\input{data/exps.tex}
\input{data/conclusion.tex}

\bibliographystyle{ACM-Reference-Format}
\bibliography{egbib.bib}

\end{document}

%% file: data/introduction.tex
\section{Introduction}
\begin{figure}[t]
	\centering
	\resizebox{0.48\textwidth}{!}{
	\includegraphics[width=\textwidth]{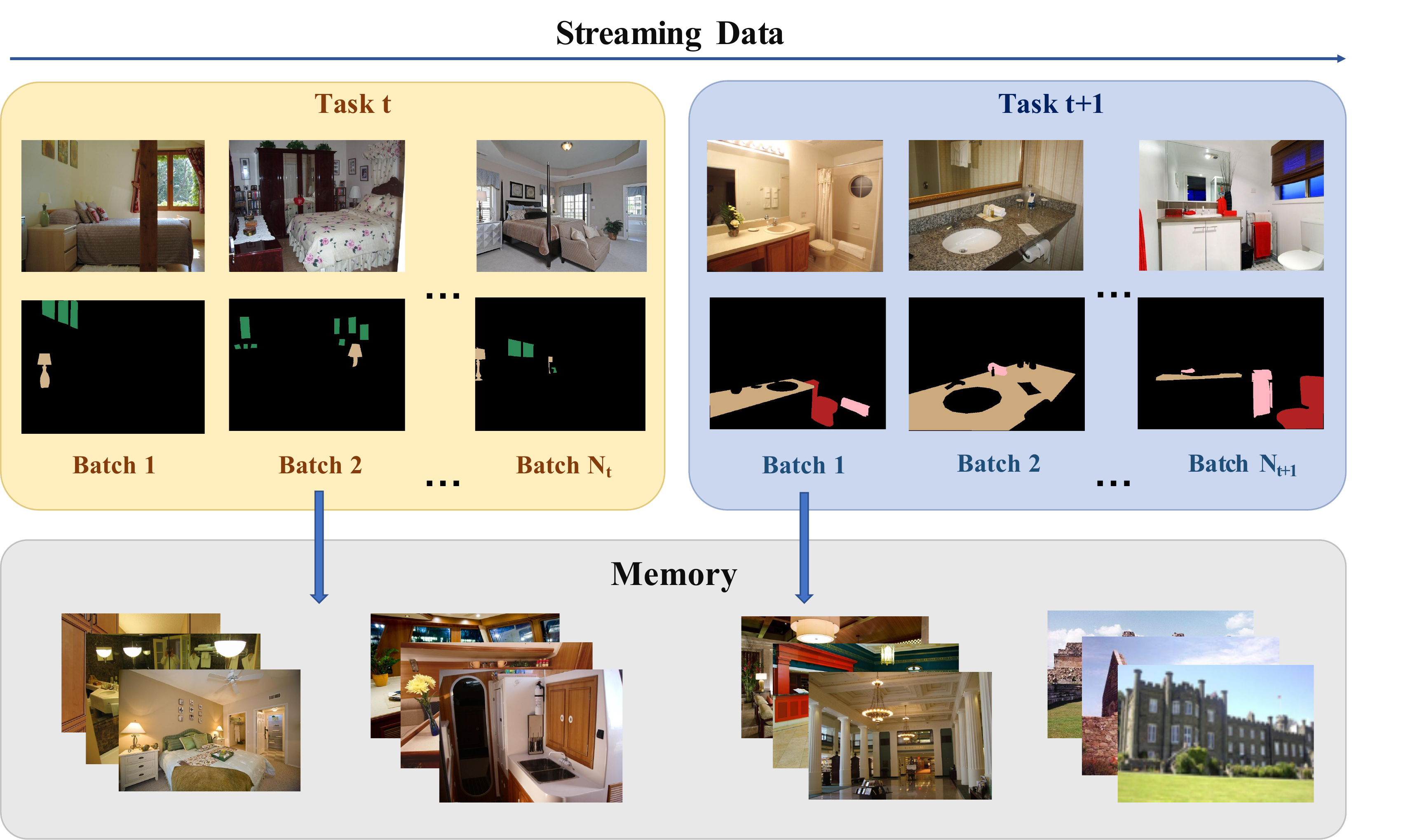}
   }
\caption{Overview of our online incremental semantic segmentation setting. Each incremental task adds images with annotations of new classes only for model adaptation, and a small set of data is saved for subsequent learning. We adopt the single-pass setting in which the data come as a stream of batches and can only be used once to update the model parameters. After each task, the model aims to segment all the semantic classes seen so far. }	\label{fig:intro}
\end{figure}
Semantic segmentation of visual scenes has recently witnessed tremendous progress  thanks to pixel-level representation learning based on deep convolutional networks. Most existing works, however, assume a close-world setting, in which all the semantic classes of interest are given when a deep segmentation network is learned. This assumption, while simplifying the task formulation, can be restrictive in real-world applications, such as medical image analysis, robot sensing, and personalized mobile Apps~\cite{li2019efficient}, where novel visual concepts need to be incorporated later on or continuously in deployment. To address this, a promising strategy is to incrementally learn semantic segmentation models~\cite{michieli2019incremental,cermelli2020modeling}.


In this work, we focus on the problem of incrementally learning semantic segmentation model from \textit{a data stream}~\cite{aljundi2019mir,aljundi2019gss}. In contrast to the offline learning, our online learning task is bounded by the run-time and only allows single-pass through the data.
Here we assume that, to reduce expensive labeling cost for semantic segmentation, we only receive annotations for the pixels of novel classes at their arrival as in \cite{cermelli2020modeling}. During training, we also assume that the system is able to 
save a subset of previously-seen training data as memory for rehearsal. This enables us to adapt the learned segmentation model to new label spaces with low annotation and training cost, and to cope with the scenarios that only limited legacy data resource can be stored due to privacy issue or data memory limitation.  
Fig.~\ref{fig:intro} illustrates the typical problem setting of this incremental semantic segmentation task. 
We note that most prior attempts on the incremental learning for semantic segmentation either adopt the offline learning setting without considering memory, or only focus on domain-specific tasks~\cite{ozdemir2018learn,tasar2019incremental,cermelli2020modeling}. By contrast, 
we aim to address the online incremental semantic segmentation with limited memory, which is more practical for many real-world scenarios. 





There are several challenges in this online class incremental semantic segmentation problem given limited memory resource. 
First, continuous learning of novel visual concepts typically involves the so-called \textit{stability-plasticity dilemma}, in which a model needs to quickly learn new concepts and meanwhile overcome catastrophic forgetting on learned visual knowledge~\cite{rebuffi2017icarl}. 
In addition, the segmentation models have to cope with \textit{background concept drift} as the background class changes due to the introduction of novel semantic classes, which further increases the difficulty of fast model adaption. 
Moreover, there also exists severe class-imbalanced problem due to varying co-occurrence of semantic classes in the data stream and small-sized memory. 

\begin{figure*}[t]
\centering
\includegraphics[width=\textwidth]{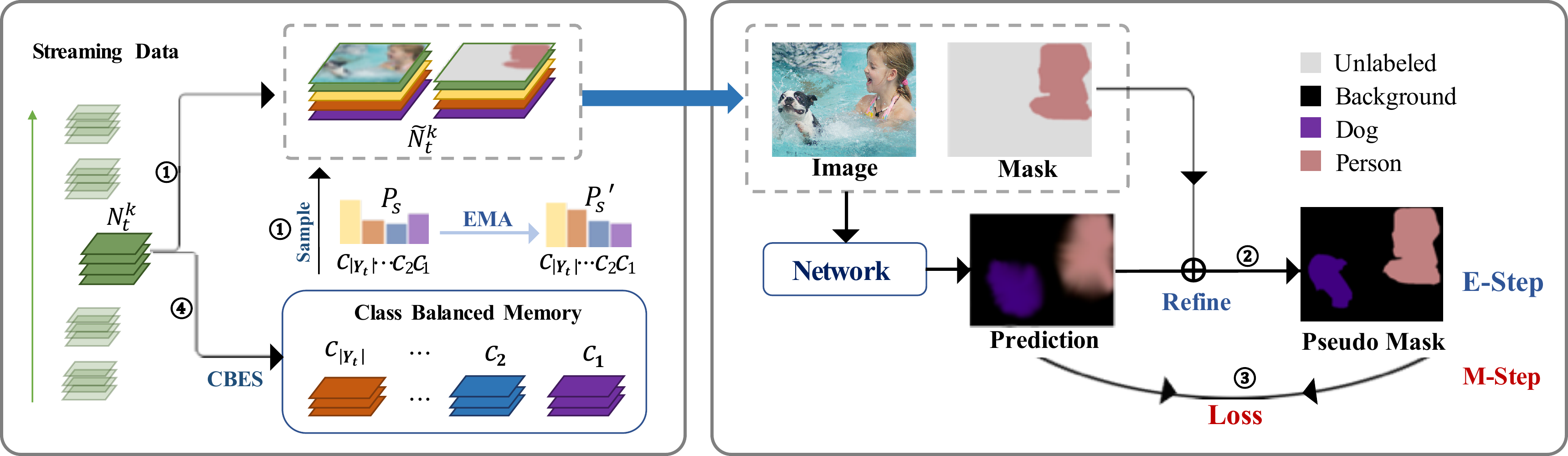}
\caption{Overview of our framework. Each incremental step begins with a new training batch $\mathcal{N}_{t}^{k}$.
	Our method first samples a batch with the same size as the new batch from a class-balancing memory with a class probability $P_{s}$, 
	which is dynamically updated at every step according to the prediction confidence of each class. 
	The sampled batch is combined with the incoming data to build the training batch.
	We then update our model based on the EM learning framework: We fill in the missing annotations using confident predictions of the trained model, which produces the pseudo-masks for the training batch at E-step. At M-step, 
	we train the model by optimizing the composite loss between the model predictions and the pseudo masks with cosine normalization.
	At the end of this incremental step, we use class-balanced exemplar selection to update the memory.
}
\label{fig:inc_algorithm}
\end{figure*}
Our goal is to tackle all three aforementioned challenges in the online incremental learning of segmentation, aiming to achieve high learning efficiency with robustness towards catastrophic forgetting and concept drift.
To this end, we propose a novel rehearsal-based deep learning strategy for building a segmentation network in an incremental manner. 
Our key idea is to train a deep network by utilizing incoming data batches at each incremental step with a \textit{dynamic sampling} policy, which concentrates on informative samples from previous steps. 
This allows us to alleviate the catastrophic forgetting problem with low training cost. 
In addition, to improve the data efficiency and overcome the background drift, we introduce a re-labeling strategy in each incremental step, which fills in the missing annotations and updates the background class using confident predictions of the trained model. 
Furthermore, we integrate two class-balancing strategies, the cosine normalization and class-balanced exemplar selection, into the network training in each step to cope with unbalanced data.

Formally, we develop a unified learning approach that integrates the above strategies into an Expectation-Maximization (EM) based learning framework. Our method starts with a deep network trained on a set of base categories, and sequentially learns novel visual classes over incremental steps. In each step, our EM learning iterates through three stages: a) sampling a small set of training data, b) filling in missing labels of pixels, and c) updating the segmentation network using the dataset with mixed true and pseudo labels. 
After the parameter update, we adopt a class-balanced reservoir sampling strategy to update the exemplar set in the memory, which is then used for the next incremental step.  

We evaluate our model on two challenging benchmarks, including PASCAL VOC 2012 and ADE20K datasets. Our empirical results and ablation study show that the proposed model achieves superior performance over prior incremental approaches, demonstrating the efficacy of our method. To summarize, the main contributions of our work are three-fold:

\begin{itemize}[leftmargin=5mm,topsep=0pt,itemsep=0pt]
	\item We introduce a new online incremental learning problem for semantic segmentation, and our approach achieves the state-of-the-art results on two challenging benchmarks. 
	\item We develop a unified EM framework that integrates a re-labeling step and a rehearsal-based training with dynamic sampling, promoting fast adaptation to novel classes and better alleviating catastrophic forgetting.
	\item To cope with imbalanced data in the online setting, we introduce cosine normalization and class-balanced reservoir sampling for incremental training of the segmentation networks. 
\end{itemize}

%% file: data/relatedwork.tex
\section{Related Work}

\paragraph{Semantic Segmentation}
Recent research has made great progress in semantic segmentation based on deep convolutional network~\cite{long2015fully, ronneberger2015u, zhao2017pyramid, ChenPKMY18deeplab}.
However, existing literature typically assumes that the semantic classes of interest and their annotated data are given in advance, which may not be feasible in practical open-world scenarios.
To address this limitation, several recent works start to explore the problem of class-incremental semantic segmentation~\cite{ozdemir2018learn,cermelli2020modeling,michieli2019incremental,tasar2019incremental}. 
In particular, ILT~\cite{michieli2019incremental} uses annotations of both novel and old classes for each incremental task and adopts knowledge distillation to alleviate forgetting.
MiB~\cite{cermelli2020modeling} revises the cross-entropy loss and knowledge distillation loss~\cite{hinton15distill} for solving forgetting and the bias caused by the background shift problem.
Notably, MiB introduces a more natural setting requiring only the pixel annotations of novel classes, which is also adopted by our work.

It is worth noting that most existing approaches focus on tackling the incremental semantic segmentation with no memory of historical training data and in an offline manner~\cite{michieli2019incremental,cermelli2020modeling}.
The only exception is the work CoRiSeg~\cite{ozdemir2018learn} for 3D medical image segmentation, which mainly addresses catastrophic forgetting with a distillation-based learning strategy on memory.
In detail, it adopts a confidence-based strategy to select stored exemplars for rehearsal in the future. 
In this work, we also consider the more practical setting as in most of incremental classification works~\cite{rebuffi2017icarl,wu2019large,yan2021dynamically} that provides a limited storage for past data.
Nevertheless, our work do not focus on the offline training for each incremental task, and aims to continuously learn the segmentation model from a non-stationary data stream in an online manner.
This enables many applications deployed on mobile devices to fast learn new concepts.

\paragraph{Class Incremental Learning}
Class incremental learning aims to learn novel concepts continuously.
There are mainly three kinds of methods in literature to solve the class incremental learning, which are regularization-based methods, distillation-based methods and structure-based methods.
Regularization-based methods~\cite{kirkpatrick2017overcoming,zenke2017continual} penalizes the change of learned parameters which are important for previously observed classes. 
Distillation-based methods~\cite{castro2018end,rebuffi2017icarl,wu2019large,douillard2020podnet} retains learned knowledge via preserving the network output on the saved exemplars with knowledge distillation.
Structured-based methods\cite{Rajasegaran19rpsnet, yan2021dynamically,abati2020conditional} separates parameters learned at different steps from each other to avoid undesirable overlapping in representations. 
For the online incremental learning setting,  most existing methods rely on the rehearsal strategy and focus on better utilization of the memory. Specifically, 
AGEM~\cite{ChaudhryRRE19agem}, which is an efficient version of GEM~\cite{lopez2017gem}, projects the gradients computed on the new data to the direction that cannot increase the loss on the data sampled from memory.
MIR~\cite{aljundi2019mir} proposes a sampling strategy selecting the maximally interfered data from memory to be learned with new data.
GSS~\cite{aljundi2019gss} focuses on the construction of memory, which aims at maximizing the sample diversity based on their gradient directions.
CBRS~\cite{ChrysakisM20cbrs} develops an class-balancing sampling strategy to deal with class-imbalanced stream of images. 
However, CBRS is designed for image classification and cannot be directly applied to semantic segmentation, which typically has instances from multiple classes in each image.
In this work, we propose a new class-balancing strategy to handle this problem.

\paragraph{Re-labeling Strategy}
Re-labeling, also known as self-training \cite{Yarowsky95self,papandreou2015weakly,hung2018adversarial}, is a strategy to learn from partially labeled or unlabeled data. It takes the most probable predictions on unlabeled data as pseudo labels for subsequent model training. 
In this work, we incorporate the idea of re-labeling into a unified EM framework in order to accommodate the background concept drift and fill in missing labels of pixels in incoming data.


%% file: data/methods.tex
\section{Our approach}\label{sec:approach}

In this section, we introduce our online incremental learning strategy for semantic segmentation, which aims to continuously learn novel visual concepts for pixel-wise semantic labeling of images. 
In particular, given a sequence of semantic classes, we consider the problem of building a segmentation model in multiple incremental tasks, each of which expands semantic label space with new arriving classes and needs to update the model with only limited data memory of previous tasks. 
In this limited-memory setting, our goal is to achieve efficient adaptation of the segmentation model without catastrophic forgetting of previously learned semantic classes.

To this end, we develop an Expectation-Maximization (EM) framework for incrementally learning a deep segmentation network. Our EM learning method incorporates a re-labeling strategy for augmenting annotations and an efficient rehearsal-based model adaptation with dynamic data sampling within a single framework.
In addition, we introduce a robust training procedure for adapting the segmentation model based on cosine normalization and class-balanced memory.
Such an integrated framework enables us to cope with the challenges of stability-plasticity dilemma, background concept drift and class-imbalance simultaneously in a principled manner.
Below we start with an introduction of our problem setting in Sec.~\ref{subsec:setup}, followed by a formal description of our model architecture in Sec.~\ref{subsec:model_arch}.
Then we present our EM-based incremental learning approach in Sec.~\ref{subsec:em_framework}.
Finally, we show our class-balancing strategies in Sec.~\ref{subsec:class_balanced}. 
\subsection{Problem Setup}\label{subsec:setup}
The online incremental learning of semantic segmentation typically starts from a segmentation model $\mathcal{S}_0$, which is trained on a base task $\mathcal{T}_0=\{\mathcal{D}^{tr}_0,\mathcal{Y}_0\}$ where $\mathcal{D}^{tr}_0$ is the training dataset, and $\mathcal{Y}_0$ are the base semantic classes. We denote the {background class} of $\mathcal{T}_0$ as $U_0$, which includes all the irrelevant classes not in $\mathcal{Y}_0$, and $\tilde{\mathcal{Y}}_0=\mathcal{Y}_0\cup \{U_0\}$ are the categories including background class at the initial task. 
The initial training set $\mathcal{D}^{tr}_0=\{(\mathbf{x}^0_i,\mathbf{y}^0_i)\}_{i=1}^{|\mathcal{D}^{tr}_0|}$, where $\mathbf{x}_i^0\in\mathbb{R}^{H\times W \times 3}$ and $\mathbf{y}_i^0\in \tilde{\mathcal{Y}}_0^{H\times W}$ are the $i$-th image and its fully-labeled annotation, respectively. 

During the incremental learning stage, the initial segmentation model receives a stream of new semantic class groups $\{\mathcal{C}_t\}_{t\geq 1}$ and their corresponding training data $\{\mathcal{D}^{tr}_t\}_{t\geq 1}$ where $t$ indicates incremental steps.
In each incremental step $t$, the training data also comes as a stream of mini-batches $\mathcal{N}^{k}_t \subset \mathcal{D}^{tr}_t$ where $k=1,\cdots,K$ and K is the number of min-batches. As in~\cite{aljundi2019mir, chaudhry2018efficient}, our online learning setting only allows the model to perform one-step parameter update using the incoming batch $\mathcal{N}^{k}_t$.  
Here we assume that only pixels from the novel classes are labeled, which may be caused by lacking previous labeling expertise and/or limited annotation budget. 
Specifically, the incoming training data $\mathcal{N}^{k}_t$ has a form of $\{(\mathbf{x}^t_i,\mathbf{y}^t_i, \mathcal{R}_i^t)\}_{i=1}^{|\mathcal{N}^{k}_t|}$ at step $t$, in which $\mathbf{y}^t_i$ is the partial annotation, and $\mathcal{R}^t_i$ is derived from $\mathbf{y}^t_i$ and indicates the labeled region in the image $\mathbf{x}^t_i$. 
For each pixel $j\in \mathcal{R}^t_i$, we have its label $\mathbf{y}^t_{i,j}\in \mathcal{C}_t$ and otherwise $\mathbf{y}^t_{i,j}$ is unlabeled. 

Moreover, we assume a small set of exemplars $\mathcal{M}_t$ from previous steps can be stored in an external memory, and the allowable size of the memory is up to $\mathcal{B}_{\mathcal{M}}$.
Typically, samples are drawn from memory, concatenated with the incoming batch and then used for training the segmentation model $\mathcal{S}_t$.  
Our goal is to learn an updated segmentation model $\mathcal{S}_t$ to incorporate the novel semantic classes in $\mathcal{C}_t$, and to achieve strong performance on a test data $\mathcal{D}^{ts}_t$. Below we refer to the model learning at each incremental step as a \textit{task}.
We note that $\mathcal{D}^{ts}_t$ includes all the semantic classes up to the $t$-th incremental step, denoted as $\mathcal{Y}_t=\mathcal{Y}_0\cup\left(\cup_{\tau=1}^t\mathcal{C}_\tau\right)$, and the corresponding \textit{background class} $U_t$.




\begin{figure*}[t]
	\begin{minipage}{0.47\linewidth}
		\centering
		\begin{algorithm}[H]
			\caption{The Method Overview.}
			\hspace*{-11.2em} {\bf Initialize:} Model $\mathcal{S}_0$; Memory $\mathcal{M}=\emptyset$
			\begin{adjustwidth}{-1em}{}
				\begin{algorithmic}
					\For{$(\mathbf{x}_i, \mathbf{y}_i, \mathcal{R}_i) \in \mathcal{D}^{tr}_0$}
						\State CBES($\mathcal{M}$, $\mathbf{x}_i$, $\mathbf{y}_i$, $\mathcal{R}_i$)
					\EndFor
					\For{\texttt{$t=1,2,3...$}}
					\While{Batch $\mathcal{N}^k_t$ comes}

					\State $\tilde{\mathcal{N}}^k_t = \mathcal{N}_t^k$
					\For{\texttt{$j=1,2,3...,|\mathcal{N}_t^k|$}}
					\State Sample the class $c_{k}$ according to $P_s(c)$ (Eq. \ref{eq:3})
					\thickmuskip=0.5\thickmuskip
					\State Denote $D_s\!=\!\{(\mathbf{x}_l,\mathbf{y}_l,\mathcal{R}_l)\in \mathcal{M}_{t}|\,\exists j\in \mathcal{R}_l,\,\mathbf{y}_{l,j}=c_k\}$
					\State Uniformly sample $(\mathbf{x}_i,\mathbf{y}_i,\mathcal{R}_i)$ from $D_s$
					\State $\tilde{\mathcal{N}}^k_t = \tilde{\mathcal{N}}^k_t \cup \{\mathbf{x}_i,\mathbf{y}_i,\mathcal{R}_i)\}$
					\EndFor
					\State Refine label $\mathbf{y}_i$ in batch $\tilde{\mathcal{N}}_t^k$ by relabeling (Eq. \ref{eq:4})

					\State Update model $\mathcal{S}$ with the loss $\mathcal{L}$ (Eq. \ref{eq:5})
					\State Update class sampling probability $P_s(c)$ (Eq. \ref{eq:3})

					\State $\mathcal{M}^{k+1}_t \leftarrow \mathcal{M}^{k}_t$ \textcolor{gray}{// Update memory}
					\For{$(\mathbf{x}_i, \mathbf{y}_i, \mathcal{R}_i) \in \mathcal{N}^{k}_t$}
					\State $\mathcal{M}^{k+1}_t$$=$$\text{CBES}(\mathcal{M}^{k+1}_{t}, \mathbf{x}_i, \mathbf{y}_i, \mathcal{R}_i)$
					\EndFor
					\EndWhile
					\State $\mathcal{M}_{t+1}^1 = \mathcal{M}_t^{k+1}$
					\EndFor
				\end{algorithmic}
			\end{adjustwidth}
			\label{alg:overview}
		\end{algorithm}
	\end{minipage}
	\hspace{5mm}
	\begin{minipage}{0.47\linewidth}
		\vspace{-12mm}

		\centering
		\begin{algorithm}[H]
			\caption{Class-balanced Exemplar Selection(CBES).}
			\hspace*{-1.50in} {\bf Input:} Memory $\mathcal{M}_t$, data $(\mathbf{x}_i,\mathbf{y}_i,\mathcal{R}_i)$
			\begin{adjustwidth}{-0.9em}{}
			\begin{algorithmic}
						
				\State Category Set $Y_i = \text{Unique}(\mathbf{y}_i) - \{U_t\}$
				\State $c_{\text{min}} = \arg \min_{c \in Y_i} |\mathcal{M}^{c}_t|$
					
				\If{ $|\mathcal{M}^{c_{\text{min}}}_t| < \mathcal{B}_{\mathcal{M}}/|\mathcal{Y}_t|$ or $|\mathcal{M}_t| < \mathcal{B}_{\mathcal{M}}$}
						\If{ $|\mathcal{M}_t| == \mathcal{B}_{\mathcal{M}}$}
						\State $c_{\text{max}} = \arg \max_{c \in \mathcal{Y}_t} |\mathcal{M}^{c}_t|$	
						\State Randomly remove an image from $\mathcal{M}^{c_{\text{max}}}_t$
						\EndIf
						\State Add $(\mathbf{x}_i, \mathbf{y}_i, \mathcal{R}_i)$ into the memory $\mathcal{M}^{c_{\text{min}}}_t$
				\Else
						\For{$c$ in $Y_i$}
						\State $m_{c}$ $\leftarrow$ the number of saved images of class $c$
						\State $n_{c}$ $\leftarrow$ the number of seen images of class $c$
						\State Sample $p\sim Uniform(0,1)$
						\If{$p \le m_{c}/n_{c}$}
							\State Randomly remove an image from $\mathcal{M}^c_t$
							\State Add $(\mathbf{x}_i, \mathbf{y}_i, \mathcal{R}_i)$ into $\mathcal{M}^c_t$
							\State Break
						\EndIf
						\EndFor
				\EndIf
			\end{algorithmic}
			\end{adjustwidth}
			\label{alg:cbes}
		\end{algorithm}
	\end{minipage}
\end{figure*}

\subsection{Model Architecture}\label{subsec:model_arch}
At each incremental step $t$, we assume a typical semantic segmentation network architecture, which consists of an encoder-decoder backbone network $\mathcal{F}_t$ extracting a dense feature map and a convolution head $\mathcal{H}_t$ that produces the segmentation score map. 
Concretely, given the image $\mathbf{x}_i$, the segmentation network first computes the convolutional feature map $\mathbf{F}_i\in\mathbb{R}^{H\times W\times D}$ using the backbone network, where $D$ is the number of channels. $\mathbf{F}_i$ is then fed into the $1\times 1$ convolution head $\mathcal{H}_t$, which has a form of linear classifier at each pixel location. Denote the parameters of the backbone network as $\mathbf{W}$, and the parameters of head as $\mathbf{\Psi} \in \mathbb{R}^{D\times(|\mathcal{Y}_t|+1)}$, our network generates its output $\mathbf{\hat{y}}_i$ as follows, 
\begin{align}
&\mathbf{F}_i=\mathcal{F}_t(\mathbf{x}_i;\mathbf{W}), \\
&p(\mathbf{\hat{y}}_i|\mathbf{x}_i)=\text{\small{Softmax}}(\mathcal{H}_t(\mathbf{F}_i))=\prod_{j\in\Omega_i}p(\mathbf{\hat{y}}_{ij}|\mathbf{F}_{ij};\mathbf{\Psi})\label{eq:model} 
\end{align}
where $\Omega_i$ represents the 2D image plane and $\mathbf{F}_{ij}\in \mathbb{R}^{D}$ for the pixel $j$. Note that our method is agnostic to specific network designs.


\subsection{Incremental Learning with EM}\label{subsec:em_framework}

We now develop an EM learning framework to efficiently train the model $\mathcal{S}_t$ with a limited memory $\mathcal{M}_t$ in an online fashion. To tackle the stability-plasticity dilemma, our strategy integrates dynamic sampling and pixel relabeling, which enables us to efficiently re-use the training data for fast model adaptation and cope with background drift caused by the partial annotation. 

Specifically, we formulate the online learning at each step as a problem of learning with latent variables and develop a stochastic EM algorithm to iteratively update the model parameters. 
In each step $t$, our learning procedure sequentially takes incoming mini-batch $\mathcal{N}_t^k$ and updates the model (denoted as $\mathcal{S}^k_t$) as well as the memory (denoted as $\mathcal{M}_t^k$). For each mini-batch, our EM algorithm iterates through three stages: 1) dynamic sampling of training data; 2) an E-step that relabels pixels with missing or background annotations, and 3) an M-step that updates the model parameters with one step of SGD. 
Below we denote the model parameter at min-batch $k$ in step $t$ as $\bm{\theta}^k_t=\{\mathbf{W}_t^k, \mathbf{\Psi}_t^k\}$ and describe the details of the three stages in each EM iteration. A complete overview of our algorithm is shown in Algorithm~\ref{alg:overview}.


\paragraph{Dynamic sampling}
We first use a rehearsal strategy~\cite{chaudhry2019tiny} to retrieve a sample of data from the memory, which has the same size as $\mathcal{N}_{t}^{k}$, and add them into the current mini-batch to build the training batch $\tilde{\mathcal{N}}_t^k$ for the subsequent EM update.
Our goal is to achieve efficient model update that learns the new classes and also to retain old ones. To this end, we develop a dynamic sampling strategy to select informative samples from the memory. 

Specifically, we devise a category-level sampling strategy that attempts to balance the training samples from the novel classes to be learned and the old ones prone to model forgetting. To achieve this, our method maintains a  \textit{class sampling probability} $P_s(c)$ for $\forall c\in \mathcal{Y}_t$, which will be described below. In order to sample a training image, we first sample a class label $c$ according to $P_s$, and then uniformly sample an image $\mathbf{x}_i$ that has pixels labeled with $c$. 
Such a sampling process allows the model to visit the data of certain classes more frequently and to improve the learning of those classes.

Our key design is to dynamically update the class probability $P_s$ according to the prediction confidence in the outputs of current model $\mathcal{S}^k_t$. 
To this end, we first compute an estimate of class confidences, denoted as $\{E(c),c\in\mathcal{Y}_t\}$, at every iteration of EM. Concretely, we use a moving average to collect the statistics from batches $\tilde{\mathcal{N}}_t^k$ as follows,
\begin{align}
E(c) &= \mu E_{{old}}(c) + (1-\mu)E_{mb}(c), \quad  \\
E_{mb}(c) &= \underset{{i\in \tilde{\mathcal{N}}^k_t, \mathbf{{y}}_{i,j}=c}}{\text{Mean}}\{p(\mathbf{\hat{y}}_{i,j}=c|\mathbf{x}_i;\bm{\theta}_t^{k-1})\}
\end{align}
where $E_{old}$ and $E_{mb}$ denotes the confidence estimate from last step and current mini-batch, respectively. Here $\mu$ is the momentum coefficient, and $\bm{\theta}_t^{k-1}$ is the model parameters after the update of last step.
At the beginning of a new task $t$, we initialize $E(c)$ as $0$ for classes in $\mathcal{C}_t$.

Given the class confidences, we define the sampling probability $P_s$ with a Gibbs distribution, which assigns higher sampling probability to the classes with lower confidence:  
\begin{align}
P_s(c) = \frac{e^{-\eta E(c)}}{\sum_{j=1}^{|\mathcal{Y}_t|} e^{-\eta E(j)}} \quad \forall c\in \mathcal{Y}_t\label{eq:3}
\end{align}
where $\eta$ is the hyperparameter to control the peakiness of the distribution.

\paragraph{E-step}

In E-step, we tackle the problem of background concept drift due to the expansion of semantic label space $\mathcal{Y}_t$, and exploit the unlabeled regions in the training data of incremental steps for model adaption. To this end, we treat the labels of the background pixels in $\{\mathcal{D}^{tr}_0\}$ and the unlabeled pixels in $\{\mathcal{D}^{tr}_t\}_{t\geq 1}$ as hidden variables, and use the current model to infer their posterior distribution given the images and semantic annotations. We then use the posterior to fill in the missing annotations for the subsequent model learning in the M-step. By augmenting the training data with those pseudo labels, we aim to further improve the training efficiency and alleviate the problem of inconsistent background annotation.     

Specifically, given a training pair $(\mathbf{x}^\tau_i, \mathbf{y}^\tau_i, \mathcal{R}_{i}^{\tau}) \in \tilde{\mathcal{N}}_t^k$, we first compute an approximate posterior of the latent part of $\mathbf{y}^\tau_i$. Here $\tau\in\{0,\cdots,t\}$ indicates the incremental step it comes from. Denote the region with latent labels as $\bar{\mathcal{R}}_i^\tau$, the posterior of its $j$-th pixel can be estimated as   
$p(\mathbf{\hat{y}}_{i,j}|\mathbf{x}^\tau_i;\bm{\theta}_t^{k-1})$ based on the current network. We then generate pseudo labels for those pixels by choosing the most likely label predictions as follows,
\begin{align}
\mathbf{{y}}^\tau_{i,j} = \underset{\mathbf{\hat{y}}_{i,j}\in\mathcal{\tilde{Y}}_t - \mathcal{C}_\tau}{\arg\max} p(\mathbf{\hat{y}}_{i,j}|\mathbf{x}^\tau_i;\bm{\theta}_t^{k-1}), \quad j\in \bar{\mathcal{R}}^i_\tau
\label{eq:4}
\end{align}
where $\mathcal{C}_\tau$ is the class group from step $\tau$. To filter out unreliable estimation, we only keep the pseudo labels of those pixels with confidence higher than a threshold $\delta$, i.e., $p(\mathbf{{y}}^\tau_{i,j}|\mathbf{x}^\tau_i;\bm{\theta}_t^{k-1})>\delta$. We adopt this hard EM to update our model in the next stage.

\paragraph{M-step}

Our M-step employs the mini-batch training data with augmented annotations to update the model parameters. In order to fully utilize the image annotations, we propose a composite loss consisting of two loss terms as follows, \begin{align}
\mathcal{L} = - \sum_{i=1}^{|\tilde{\mathcal{N}}^k_t|} \left(\sum_{j=1}^{|\mathbf{{y}}^\tau_{i}|} \log p(\mathbf{{y}}^{\tau}_{i,j}|\mathbf{x}_i^\tau;\bm{\theta}_t^k) 
	+ \gamma\sum_{j=1}^{|\bar{\mathcal{R}}^\tau_i|} \log \sum_{z \in \tilde{\mathcal{Y}}_{t}\backslash \mathcal{C}_\tau} p( z|\mathbf{x}_i^\tau;\bm{\theta}_t^k)\right) 
\label{eq:5}
\end{align} 
where $p(\cdot)$ is defined as in Equation~\eqref{eq:model} and $\gamma$ is a weighting coefficient. The first term is the standard cross-entropy loss on every pixel with real annotation or pseudo labels, while the second term penalizes the model on classifying the unlabeled pixels at step $\tau$ into label space $\mathcal{C}_\tau$, which encourages the labeling consistent with the partial groundtruth annotations.

\subsection{Class Balancing Strategies}\label{subsec:class_balanced}
The online incremental learning of semantic segmentation typically has to face severe class imbalance in each step, which is caused by the limited mini-batch size and the data sampling. To address this issue, we introduce two strategies into the training of the segmentation network, which are detailed below.   


\paragraph{Cosine Normalization} The model output at each pixel $j$ is represented as a probability vector $p(\mathbf{\hat{y}}_{ij}|\mathbf{F}_{ij};\mathbf{\Psi}^k_t)$ with dimension $|\tilde{\mathcal{Y}}_t|$. Here the $m$-th element of the probability is defined as follows
\begin{align}
	p(\mathbf{\hat{y}}^m_{ij}|\mathbf{F}_{ij};\mathbf{\Psi}^k_t) = \frac{e^{\tau d(\mathbf{u}_m , \mathbf{F}_{ij})}}{\sum_{c=1}^{|\tilde{\mathcal{Y}}_t|} e^{\tau d(\mathbf{u}_c , \mathbf{F}_{ij})}}
\end{align}
where $\tau$ is the temperature used to control the sharpness of the softmax distribution, $\mathbf{u}_m\in\mathbb{R}^D$ is the m-th column of classifier weight $\mathbf{\Psi}^k_t$, and $d$ denotes the cosine distance. This enables us to alleviate the influence of biased norm due to class imbalance and produces more balanced scores across categories.
It is worth noting that we are the first to adopt cosine normalization in the online class incremental learning problem, even though it has been used in other problems~\cite{luo2018cosine,qi2018low, zhang2021distribution}.

\paragraph{Class-balanced Exemplar Selection} 
We then extend the class-balanced reservoir sampling (CBRS)~\cite{ChrysakisM20cbrs} to the incremental learning of semantic segmentation by taking into account the multiple classes in each image.
Specifically, at the step $t$, we represent the memory for class $c$ as $\mathcal{M}_t^c = \{(\mathbf{x}_i,\mathbf{y}_i, \mathcal{R}_i)|c \in \text{Unique}(\mathbf{y}_i)\backslash\{U_t\}\}$.
Here we ignore the mini-batch index $k$ for notation clarity.
We build a class-balanced memory by maximizing the minimum size of $\mathcal{M}_t^c$, denoted as $\mathcal{M}_t^{c_{\text{min}}}$.
Concretely, we save an image with the class $c_{\text{min}}$ into the memory if  $|\mathcal{M}_t| < \mathcal{B}_{\mathcal{M}}$ or $\mathcal{M}_t^{c_{\text{min}}}$ is below the average size, i.e., $\mathcal{M}^{c_{\text{min}}}_t$ is smaller than $\mathcal{B}_{\mathcal{M}}/|\mathcal{Y}_t|$.
Otherwise, we perform reservoir sampling for all the classes in $\mathbf{y}^t_i$ until an image is selected.
Algorithm~\ref{alg:cbes} describes the details of exemplar selection strategy.



%% file: data/exps.tex
\section{Experiments}\label{sec:exps}

We evaluate our method\footnote{Code is available at \url{https://github.com/Rhyssiyan/Online.Inc.Seg-Pytorch}} on two online incremental learning benchmarks of semantic segmentation, which are built on the PASCAL VOC 2012~\cite{everingham2015pascal} and ADE20K~\cite{zhou2017scene} datasets, respectively. 
Below we first introduce our experimental setup in Sec.~\ref{subsec:exp_setup}, followed by reporting our results and analysis on the PASCAL VOC 2012 benchmark in Sec.~\ref{subsec:voc_exp} and ADE20K in Sec.~\ref{subsec:ade_exp}. Finally, we conduct an ablation study on the ADE20K dataset to analyze the contribution of our method components in Sec.~\ref{subsec:abl_study}.

\vspace{-1mm}
\subsection{Experiment Setup}\label{subsec:exp_setup}
We now describe the setup of our experimental evaluation from three aspects, including the learning protocol of deep segmentation networks, baseline methods for comparison and evaluation metrics.  
\begin{table*}[t]
	\centering
	\setlength{\tabcolsep}{3pt} 
	\caption{Mean IoU on the Pascal-VOC 2012 dataset for different incremental semantic segmentation splits.  $*$ means the original method does not use an exemplar set and we apply the method on both incoming data and memory. The experiments are conducted under memory size $\mathcal{B}_\mathcal{M}=$100.}	\resizebox{0.99\textwidth}{!}{
	\begin{tabular}{l||cc|c||cc|c||cc|c||cc|c||cc|c||cc|c}
		\toprule
		\multirow{3}{*}{Methods}    & \multicolumn{6}{c||}{\textbf{{19-1}}}   & \multicolumn{6}{c||}{{\textbf{15-5}}} & \multicolumn{6}{c}{{\textbf{15-1}}}    \\
		    & \multicolumn{3}{c||}{\bf{Disjoint}}        & \multicolumn{3}{c||}{\bf{Overlapped}}  & \multicolumn{3}{c||}{\bf{Disjoint}}     & \multicolumn{3}{c||}{\bf{Overlapped}}  & \multicolumn{3}{c||}{\textbf{Disjoint}}      & \multicolumn{3}{c}{\textbf{Overlapped}} \\
		 & \it{0-19}  & \it{20}   & \it{all}  & \it{0-19}  & \it{20}   & \it{all}     & \it{0-15}  & \it{16-20}   & \it{all}   & \it{0-15}  & \it{16-20}   & \it{all} & \it{0-15}  & \it{16-20}   & \it{all}  & \it{0-15}  & \it{16-20}   & \it{all}     \\ 
		\midrule
		{ER~\cite{chaudhry2019tiny} }        & $71.04$ & $23.36$ & $68.77$ & $71.02$ & $15.96$ & $68.39$ & $65.20$ & $28.94$ & $56.57$ & $65.37$ & $27.84$ & $56.44$ & $66.6$ & $24.98$  & $56.69$ & $68.64$ & $15.37$ & $55.95$ \\
		{LwF.MC*~\cite{rebuffi2017icarl} }   & $72.65$ & $27.56$ & $70.51$ & $72.12$ & $24.46$ & $69.85$ & $69.89$ & $36.55$ & $61.95$ & $71.85$ & $36.26$ & $63.38$ & $68.68$ & $30.81$ & $59.66$ & $69.73$ & $30.19$ & $60.32$ \\
		{iCaRL~\cite{rebuffi2017icarl} }     & $58.81$ & $8.73$ & $56.44$ & $60.38$ & $7.63$  & $57.87$ & $60.06$ & $19.71$ & $50.45$ & $59.39$ & $24.97$ & $51.19$ & $61.48$ & $3.23$  & $47.61$ & $63.78$ & $7.35$  & $50.35$ \\
		{AGEM~\cite{chaudhry2018efficient} } & $71.44$ & $10.17$ & $68.52$ & $72.19$ & $9.11$ & $69.19$ & $64.09$ & $16.83$ & $52.83$ & $65.74$ & $16.91$ & $54.12$ & $66.90$ & $19.67$  & $55.66$ & $65.83$ & $20.18$  & $54.96$ \\
		{MIR~\cite{aljundi2019mir} }         & $68.01$ & $17.57$ & $65.61$ & $67.80$ & $13.68$ & $65.22$ & $64.57$ & $29.00$ & $56.10$ & $64.31$ & $28.28$ & $55.73$ & $64.46$ & $3.36$  & $52.41$ & $67.94$ & $4.67$  & $52.87$ \\
		{MiB*~\cite{cermelli2020modeling} }  & $68.01$ & $27.69$ & $69.92$ & $71.97$ & $21.89$ & $69.59$ & $69.62$ & $34.87$ & $61.35$ & $70.25$ & $32.77$ & $61.33$ & $69.37$ & $29.1$ & $59.78$ & $69.45$ & $29.45$  & $59.92$ \\
		{CoRiSeg~\cite{ozdemir2018learn} }   & $72.19$ & $24.10$ & $69.90$ & $72.90$ & $23.16$ & $70.53$ & $70.49$ & $37.43$ & $62.62$ & $73.93$ & $44.37$ & $66.89$ & $70.73$ & $13.35$ & $57.07$ & $72.56$ & $13.32$ & $58.45$ \\
		\midrule
		{Ours }                              & $\bm{73.43}$ & $\bm{39.10}$ & $\bm{71.80}$ & $\bm{73.76}$ & $\bm{43.42}$ & $\bm{72.32}$ & $\bm{73.45}$ & $\bm{47.89}$ & $\bm{67.36}$ & $\bm{75.56}$ & $\bm{49.89}$ & $\bm{69.45}$ & $\bm{73.88}$ & $\bm{37.15}$ & $\bm{65.14}$ & $\bm{75.77}$ & $\bm{40.34}$ &  $\bm{67.33}$ \\
		\bottomrule
	\end{tabular}}
	\label{tab:pascal_voc_splits}
\end{table*}

\begin{table*}[t]
	\centering
	\caption{The performance of various methods with the fixed memory size $\mathcal{B}_{\mathcal{M}} \in \{20, 50, 100\}$, respectively. The experiments are conducted on the 15-1 disjoint split of PASCAL VOC 2012 dataset. $*$ means the original method does not use an exemplar set and we apply the method on both incoming data and memory}
	\resizebox{0.65\textwidth}{!}{
		\footnotesize
		\begin{tabular}{l|cc|cc|cc}
			\toprule[0.3mm]
			\multirow{2}{*}{\textbf{Methods}}         & \multicolumn{2}{c|}{$\mathcal{B}_{\mathcal{M}}=20$}   & \multicolumn{2}{c|}{$\mathcal{B}_{\mathcal{M}}=50$}             & \multicolumn{2}{c}{$\mathcal{B}_{\mathcal{M}}=100$}                                                                                                                          \\
				\cmidrule{2-7}
			& \textbf{imIoU}                     & \textbf{Final mIoU}                                      & \textbf{imIoU}                     & \textbf{Final mIoU}                                       & \textbf{imIoU}                     & \textbf{Final mIoU}                                       \\
			\midrule
		ER~\cite{chaudhry2019tiny}  & $44.75$ & $34.57$ & $55.30$ & $49.45$ & $61.70$ & $56.69$ \\
		LwF.MC*~\cite{rebuffi2017icarl}	 & $54.46$  & $45.91$  & $60.20$ & $53.25$ & $63.74$ & $59.66$ \\
		iCaRL~\cite{rebuffi2017icarl} & $37.93$ & $30.43$  & $49.07$ & $42.74$ & $52.96$ & $47.61$ \\
		AGEM~\cite{chaudhry2018efficient} &  $50.84$  & $43.89$  & $52.80$ & $46.98$ & $54.15$ & $51.01$ \\
		MIR~\cite{aljundi2019mir} & $47.07$ & $37.46$  & $54.83$ & $47.14$ & $61.81$ & $52.41$ \\
		MiB*~\cite{cermelli2020modeling} & $56.98$  & $48.90$  & $61.93$ & $54.34$ & $64.41$ & $59.78$ \\
		CoRiSeg~\cite{ozdemir2018learn} & $54.50$ & $45.42$ & $60.73$ & $52.06$ & $64.59$ & $57.07$ \\
		\midrule
		Ours	 & $\bm{64.64}$  & $\bm{60.49}$  & $\bm{65.57}$ & $\bm{60.14}$ & $\bm{68.06}$ & $\bm{65.14}$ \\
		\bottomrule[0.3mm]
	\end{tabular}}
	\label{tab:pascal_voc_mem_size}
\end{table*}
\vspace{-1mm}
\paragraph{Online class-incremental Learning Protocol}
To evaluate the performance of online incremental learning methods on fast adaptation and catastrophic forgetting, we adopt the following training protocol. 
Following the dataset split proposed in~\cite{cermelli2020modeling}, we start from an initial model trained on a set of base classes and divide the remaining classes into different groups, which defines the incremental tasks to be learned sequentially.
The word "online" means data are coming in streaming form, i.e., each data batch can be seen only once.
In addition, it is allowed to save limited history data in a memory buffer of fixed size.



\paragraph{Comparison Methods} 
To demonstrate the efficacy of our framework, we choose a set of typical incremental learning methods as our comparisons. 
Experience replay (ER)~\cite{chaudhry2019tiny} is used as a basic continual learning strategy.
LWF~\cite{li2017learning} and ICARL~\cite{rebuffi2017icarl} are two widely adopted methods designed for the offline class-incremental classification.
AGEM~\cite{ChaudhryRRE19agem} and MIR~\cite{aljundi2019mir} are the methods for the online class-incremental classification.
MIB~\cite{cermelli2020modeling} and CoRiSeg~\cite{ozdemir2018learn} are two offline incremental learning methods for semantic segmentation but only the latter considers utilizing memory data.



\paragraph{Evaluation Metrics}
We evaluate the model performance at the end of each incremental task.
For each task, we use the mean of class-wise intersection over union (mIoU) as our metric, which includes the background class at each task.
In addition, we take the \textit{incremental mean IoU} (imIoU), which is computed as the average of the mIoU over different tasks, to measure the overall performance of the model over time.

%

\begin{table*}[t]
	\centering
	\setlength{\tabcolsep}{6pt} 
	\caption{Mean IoU on the ADE20K dataset for different class incremental learning scenarios. $*$ means the original method does not use an exemplar set and we apply the method on both incoming data and memory.}
	\resizebox{0.99\textwidth}{!}{
	\begin{tabular}{l||cc|c||cccccc|c||ccc|c}
		\toprule
		\multirow{2}{*}{Methods} & \multicolumn{3}{c||}{{\textbf{100-50}}} & \multicolumn{7}{c||}{{\textbf{100-10}}} & \multicolumn{4}{c}{{\textbf{50-50}}} \\
		\cmidrule{2-15}
		   & \textit{0-100} & \textit{101-150} & \textit{all}  & \textit{0-100} & \textit{100-110} & \textit{110-120} & \textit{120-130} & \textit{130-140} & \textit{140-150} & \textit{all}  & \textit{0-50} & \textit{51-100} & \textit{101-150} & \textit{all}  \\ \hline
		ER~\cite{chaudhry2019tiny}        & $33.02$ & $2.71$ & $22.98$ & $32.59$ & $3.48$ & $4.17$ & $0.01$ & $\bm{1.61}$ & $0.08$ & $22.42$ & $31.70$ & $7.57$ & $3.22$ & $14.28$ \\
		LwF.MC*~\cite{rebuffi2017icarl}   & $34.71$ & $3.69$ & $24.44$ & $33.09$ & $4.65$ & $4.14$ & $0.13$ & $1.60$ & $\bm{0.20}$ & $22.84$ & $34.99$ & $9.07$ & $4.14$ & $16.19$  \\
		iCaRL~\cite{rebuffi2017icarl}     & $21.63$ & $0.00$ & $14.46$ & $22.77$ & $0.00$ & $0.00$ & $0.00$ & $0.00$ & $0.01$ & $15.22$ & $28.13$ & $0.00$ & $0.00$ & $9.50$ \\
		AGEM~\cite{chaudhry2018efficient} & $32.68$ & $0.05$ & $21.88$ & $31.50$ & $0.20$ & $1.35$ & $0.00$ & $0.14$ & $0.00$ & $21.18$ & $31.15$ & $2.16$ & $0.00$ & $11.23$ \\
		MIR~\cite{aljundi2019mir}         & $35.61$ & $0.00$ & $23.82$ & $35.31$ & $0.00$ & $0.00$ & $0.00$ & $0.00$ & $0.00$ & $23.62$ & $37.18$ & $0.91$ & $0.00$ & $12.86$ \\
		MiB*~\cite{cermelli2020modeling}  & $34.01$ & $3.22$ & $23.81$ & $32.11$ & $\bm{5.12}$ & $3.71$ & $0.01$ & $1.41$ & $0.13$ & $22.17$ & $33.55$ & $8.61$ & $3.58$ & $15.37$ \\
		CoRiSeg~\cite{ozdemir2018learn}   & $36.58$ & $0.36$ & $24.58$ & $32.41$ & $0.00$ & $0.21$ & $0.00$ & $0.00$ & $0.00$ & $21.69$& $40.20$ & $1.25$ & $0.31$ & $14.10$ \\

		\midrule
		Ours                              & $\bm{36.80}$ & $\bm{5.33}$ & $\bm{26.38}$ & $\bm{36.66}$ & $4.83$ & $\bm{12.87}$ & $\bm{1.25}$ & $0.42$ & $0.06$ & $\bm{25.81}$ & $34.44$ & $\bm{15.22}$ & $\bm{7.15}$ & $\bm{19.04}$ \\ 
		\bottomrule
	\end{tabular}}
	\label{tab:ade_mem750}
\end{table*}
\subsection{PASCAL VOC 2012}\label{subsec:voc_exp}

\paragraph{Dataset} The PASCAL VOC 2012 dataset~\cite{everingham2015pascal} consists of 10582 images in the training split and 1449 images in the validation split with a total of 21 different classes, including a background class given by the dataset. 
We use the validation split as our test set as the original test set is not publicly available.
Following MiB\cite{cermelli2020modeling}, we also evaluate our method on both \textit{disjoint} and \textit{overlapped} setup where disjoint setup means that the images of each task is disjoint, and overlapped means there exist overlapped images between tasks.
Moreover, we test our method on three different splits on both disjoint and overlapped setups.
We begin with a model trained on 15 foreground classes, and the novel classes come incrementally in two streaming patterns, including 5 classes added at once (15-5) and sequentially, 1 at a time (15-1).
Besides, we also start from a model trained on 19 foreground classes and one class is added at once (19-1).
For the evaluation, we build a test set by taking all the images that contains known classes from the validation split.

\begin{figure*}[t]
	\centering
	\includegraphics[width=\textwidth]{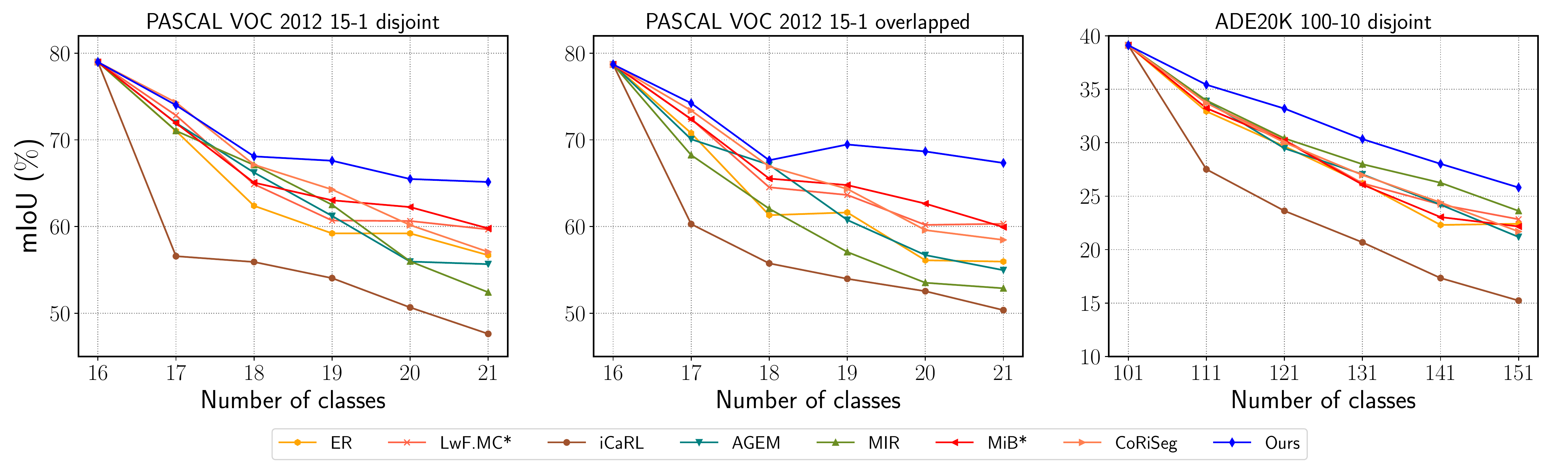}
	\caption{The mIoU of various methods with the Deeplab-v3 at different incremental tasks. The left two figures are the experiments on the VOC 2012 benchmark and the right two figures are on the ADE20K benchmark.}
	\label{fig:voc_steps_perf}
\end{figure*} 
\paragraph{Implementation Details} 
We conduct our experiments based on DeepLab-v3~\cite{chen2017rethinking} with the ResNet101~\cite{he2016deep} as the backbone.
Following the protocol used in online incremental classification~\cite{aljundi2019mir}, we hold out a validation set from the original training data to tune the hyper-parameters.
The hyper-parameters of our method are $\delta=0.8, \gamma=0.5, \eta=1, \mu=0.9, \tau=12$.
At the initial step, we start with the model pretrained on ImageNet and train the model for 60 epochs and adopt the SGD optimizer with batch size 24, momentum 0.9, weight decay 1e-4. 
The learning rate decays following the polynomial decay rule with power $0.9$. 
For the incremental stage, the incoming data comes as mini-batches with batch size 4. 
We concatenate the incoming batch with a same-sized batch(size=4) sampled from memory to form our mini-batch(size=8) for updating the model.
We adopt SGD optimizer with learning rate 0.001 and momentum 0.9.

\paragraph{Quantitative Results}
Tab.~\ref{tab:pascal_voc_splits} summarizes the comparison results on PASCAL VOC 2012 benchmark.
We can see that our method consistently outperforms other methods by a sizable margin at different incremental splits.
Moreover, our method achieves better performance both on the old classes and the novel classes, which demonstrates that our strategy achieves faster adaptation on the streaming setting and is more robust towards catastrophic forgetting.
Specifically, under the disjoint split of 15-1 setting, compared with the best baseline MiB*, we improve the final mIoU from  59.78 to 65.14(+\textbf{5.36}).
We also plot the curves of mIoU at each incremental task for different methods in Fig.~\ref{fig:voc_steps_perf}.
We can see that the gap between our method and baselines increases over time.

\paragraph{Effects of Memory Size} We also conduct extensive experiments on the 15-1 disjoint split to explore the effect of memory size. 
The memory size(i.e. number of saved exemplars) ranges from 20 to 100, which is equivalently 1 to 5 exemplars per class on average, indicating different difficulty of the scenario.
As shown in Tab.~\ref{tab:pascal_voc_mem_size}, our method consistently outperforms other methods on three memory sizes, for example, surpassing MiB by $7.66$ imIoU when $\mathcal{B}_\mathcal{M}=20$.


\begin{table}[t]
	\centering
	\footnotesize
	\caption{Ablation Study on 50-50 split of ADE20K dataset. \textbf{D.S.}: dynamic sampling, \textbf{C.R.}: composite loss with relabeling, \textbf{C.N.}:  cosine normalization, \textbf{CBES}: class-balanced exemplar selection.}
		\resizebox{0.35\textwidth}{!}{
			\begin{tabular}{cccc|c}
				\toprule[0.28mm]
				\multicolumn{4}{c}{Components}   & \multirow{2}{*}{imIoU(\%)}  \\
				CBES & C.R. &  C.N. & D.S. &    \\
				\midrule
				\xmark &\xmark & \xmark & \xmark & $17.94$  \\
				\checkmark & \xmark & \xmark & \xmark & $18.63$ \\
				\checkmark & \checkmark &  \xmark & \xmark &  $19.32$  \\
				\checkmark & \checkmark &  \checkmark  & \xmark &  $21.18$  \\
				\checkmark & \checkmark & \checkmark & \checkmark & $22.85$ \\
				\bottomrule[0.28mm]
			\end{tabular}
	}
	\label{tab:abl_ade}
	\vspace{-3mm}
\end{table}

\subsection{ADE20K}\label{subsec:ade_exp}
\paragraph{Dataset} ADE20K~\cite{zhou2017scene} is a large dataset for scene segmentation with 151 classes, including a background class given by the dataset. 
The dataset contains around 20K training, 2K validation and 3K test images.
We evaluate our algorithm on the validation set as the test set has not been released.
Like MiB~\cite{cermelli2020modeling}, we adopt disjoint setup for ADE20K by spliting training images into disjoint image sets.
Specifically, we begin with a model pretrained on 100 foreground classes and incrementally add the remaining 50 classes at once(100-50), or sequentially, 10 at a time(100-10).
We also start from a model pretrained on 50 foreground classes and then add the remaining 100 classes sequentially, 50 at a time, denoted as 50-50.
Besides, we also start from a model pretrained on 50 foreground classes and fifty classes are added sequentially, denoted as 50-50.
The evaluation is consistent with PASCAL VOC 2012 benchmark, which builds the test set at task $t$ by selecting all images of known classes from validation set.


\paragraph{Implementation Details}
At step 0, we start with the model pretrained on the ImageNet Dataset and train the model for 60 epochs and adopt the SGD optimizer with batch size 24, momentum 0.9, weight decay 1e-4. The learning rate starts with 1e-2 and decays following the polynomial decay rule with power 0.9.
The hyper-parameters of our method are $\delta=0.8, \gamma=0.1, \eta=10, \mu=0.9, \tau=12$.
For each incremental task, we start with constant learning rate 1e-3 , where the batch size is 8, weight decay 0.0001.
Specifically, both the size of incoming mini-batch and the mini-batch retrieved from memory are 4.
The other configurations are kept same as the ones in the experiments of PASCAL VOC 2012.

\begin{figure}[t]
	\centering
	\begin{minipage}{0.5\textwidth}
		\centering
		\includegraphics[width=0.95\textwidth]{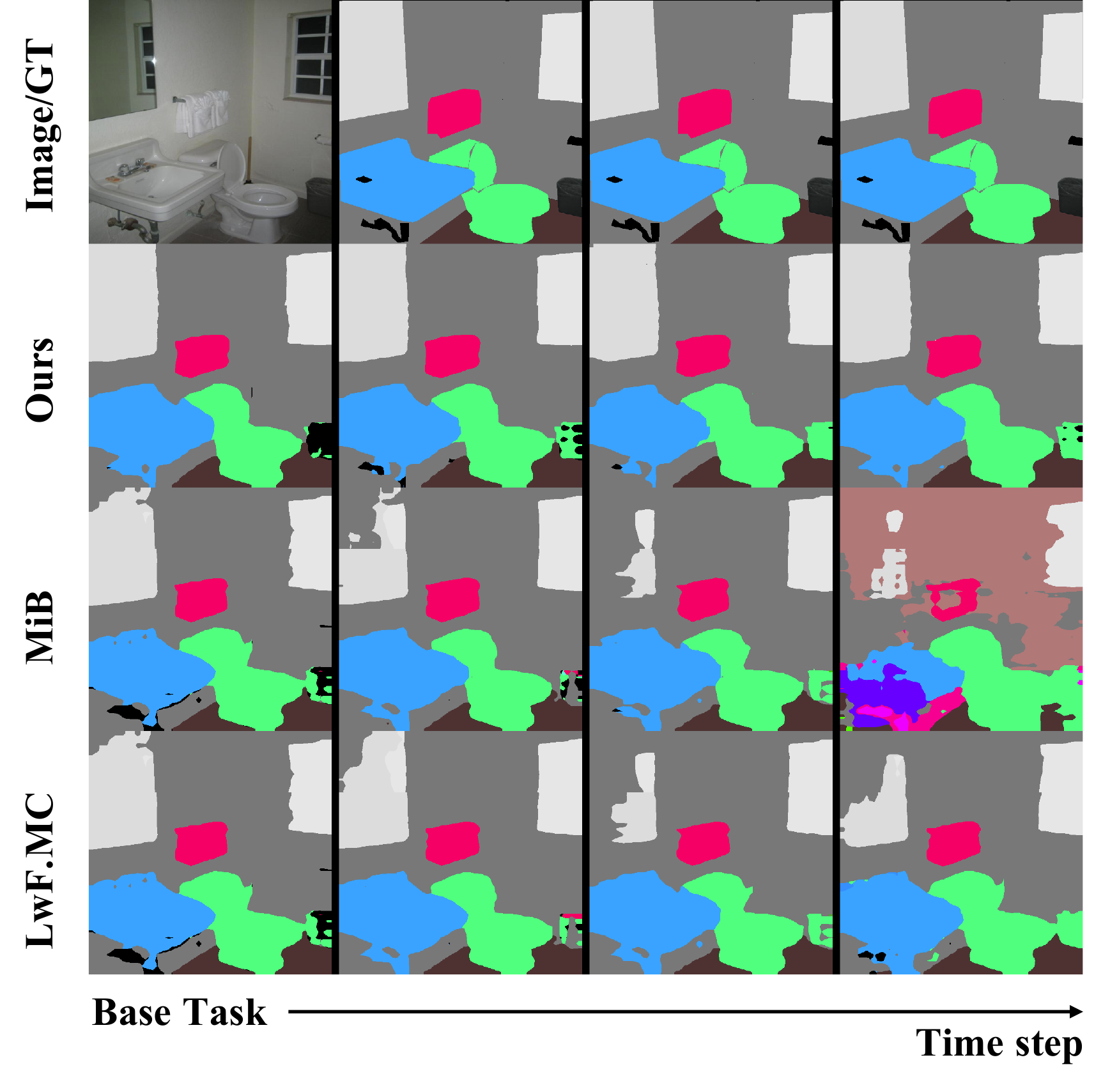}
	\end{minipage}
	\caption{Visualization of our method, MiB and LwF.MC on ADE20K dataset. Top row shows the image and the change of actual ground truth over steps. The bottom three rows present the prediction of our method, MiB and LwF.MC at different tasks.}\vspace{-2mm}
	\label{fig:visual}
\end{figure}

\paragraph{Quantitative Results}
In Tab.~\ref{tab:ade_mem750}, we compare our method with previous methods on ADE dataset.
It is evident that our approach still consistently surpasses other methods with a large margin. 
Specifically, we outperform LWF.MC* with $2.85\%$ mIoU on the 50-50 split.
As shown in Fig.~\ref{fig:voc_steps_perf}, our method achieves more than 3$\%$ imIoU improvement at final task.

\paragraph{Qualitative Comparison}
Fig.~\ref{fig:visual} shows visualization of the results of our method and two other methods. We can see that our method are more robust towards catastrophic forgetting compared with other methods.

\vspace{-1mm}
\subsection{Ablation Study}\label{subsec:abl_study}
In Tab.~\ref{tab:abl_ade}, we conduct a series of ablation studies on ADE20K to evaluate the effect of our model components.
We can see that each component plays a significant role in improving the final performance.
In particular, class-balanced exemplar selection strategy outperforms the baseline ER by $0.69$ on imIoU.
After adding the composite loss, the performance is improved from $18.63$ to $19.32$.
Moreover, the cosine normalization further brings up the  imIoU by $1.55$.
In the end, the dynamic sampling strategy boosts the method from $21.18$ to $22.85(+\bm{1.67})$.

%% file: data/conclusion.tex
\section{Conclusion}\label{sec:conclusion}
In this paper, we have introduced an online incremental learning problem for semantic segmentation, which is more practical for real-world applications. Our problem setting assume that a limited memory is available for model learning while the data comes in a streaming manner and the model updates its parameters only once for each incoming batch.
To address this challenging problem, we develop a unified EM learning framework that integrates a re-labeling strategy for missing pixel annotations and an efficient rehearsal-based incremental learning step with dynamical sampling.
Moreover, we also introduce cosine normalization and class-balanced memory to solve the class-imbalanced problem.
Extensive experimental results on the PASCAL VOC 2012 and the ADE20K dataset demonstrate the advantages of our methods.